\documentclass[aps, prb, showpacs, twocolumn, amsmath, letterpaper]{revtex4-2}

\usepackage{graphics}
\usepackage{graphicx}
\usepackage{booktabs}

\usepackage{amssymb}
\usepackage{bm}

\usepackage[english]{babel}

\usepackage{tabularx}
\usepackage{booktabs}
 
\usepackage{amsmath}
\usepackage{graphicx}
\usepackage{wrapfig}

\usepackage{adjustbox}
\usepackage{mathtools}  

\usepackage[colorlinks=true, allcolors=blue]{hyperref}

\begin{document}

\title{Endogenous Regime Switching Driven by Scalar-Irreducible Learning Dynamics}

\author{Sheng Ran}

\affiliation{Department of Physics, Washington University in St. Louis, St. Louis, MO 63130, USA}

\affiliation{Reconstructing Future Research Initiative}

\date{\today}

\begin{abstract}

Achieving endogenous regime switching is crucial for the emergence of autonomous intelligence, yet remains a central challenge for existing machine learning frameworks, where such transitions are typically externally imposed. In this work, we introduce a classification that distinguishes scalar-reducible dynamics, which can be expressed as gradient flows driven by a scalar objective, from scalar-irreducible dynamics that cannot be reduced to such a form. While most existing machine learning systems operate within the scalar-reducible class, we demonstrate that scalar-irreducible dynamics naturally enable internally generated regime switching through feedback between fast dynamical variables and slow structural adaptation. Using a minimal dynamical model, we illustrate how this mechanism produces sustained endogenous regime transitions without external scheduling. Our results suggest a new dynamical paradigm for regime exploration and provide a potential route toward autonomous learning systems whose adaptive behavior is organized internally rather than externally prescribed.

\end{abstract}

\maketitle{}

\section{Introduction}

Machine learning systems can be naturally viewed as dynamical processes that evolve in high-dimensional state or parameter spaces~\cite{E2017,Su2016}. During learning, the system often traverses different dynamical regimes, such as quiescent states, oscillatory behaviors, or phases of rapid structural reorganization~\cite{Lewkowycz2020,Chaudhari2018}. The ability to switch between such regimes can be crucial for effective learning, since remaining confined within a single regime may limit exploration and prevent the system from discovering improved configurations~\cite{Dauphin2014}.

The significance of regime switching becomes even more fundamental in autonomous learning. Unlike conventional machine learning systems, which can rely on externally imposed schedules or objectives to trigger changes in behavior, an autonomous learner must regulate such transitions from within. 
It must decide, through its own dynamics, whether to remain in a given mode of operation or to enter a different regime in order to explore, restructure, or adapt~\cite{Oudeyer2007,Schmidhuber2010}. For this reason, regime switching is not merely an auxiliary feature of autonomous learning but a necessary condition for it. A system locked into a single regime may still optimize locally, but it cannot change the manner in which it learns when that mode becomes inadequate. The ability to transition between regimes is therefore a prerequisite for self-regulated exploration and, in a deeper sense, for the emergence of autonomous intelligence.

In current machine learning practice, regime transitions are typically induced through external mechanisms. Examples include learning-rate schedules, annealing procedures~\cite{Loshchilov2016}, noise injection~\cite{Neelakantan2015}, curriculum learning~\cite{Bengio2009}, or other forms of externally imposed parameter modulation. Although effective in many contexts, these approaches rely on external control and therefore do not provide a mechanism for regime switching that arises intrinsically from the learning dynamics.

In this work we propose a different approach, in which regime switching emerges endogenously from the internal dynamics of the learning system. Our approach is motivated by a structural classification of learning dynamics based on whether the dynamics can be reduced to a gradient flow driven by a scalar potential. Under this perspective, learning dynamics can be divided into two broad classes: \emph{scalar-reducible dynamics}, which can be expressed as gradient-driven flows of a scalar objective, and \emph{scalar-irreducible dynamics}, which cannot be reduced to such a form.

Our analysis indicates that most existing machine learning systems fall into the scalar-reducible class, where the dynamics are effectively driven by the gradient of a scalar objective function. 
We argue that this structural restriction strongly limits the system's ability to generate sustained endogenous regime transitions. By contrast, scalar-irreducible dynamics naturally enable internally generated regime switching through feedback between fast dynamical variables and slow structural adaptation. To illustrate this mechanism, we construct a minimal dynamical learning model that incorporates scalar-irreducible structure. We show that this system exhibits repeated, internally generated regime transitions without relying on externally imposed schedules. 

The scalar-irreducible mechanism proposed here is largely absent from current machine learning architectures. 
Beyond providing a new route for endogenous regime switching, it suggests a broader alternative paradigm for designing learning systems, particularly in the context of autonomous learning where internal dynamical organization must replace externally imposed objectives or schedules.

\section{Scalar-Reducible vs Scalar-Irreducible Dynamics}

Learning algorithms can be naturally formulated as dynamical systems evolving in parameter space. Consider a model with parameters $x \in \mathcal{M}$, where $\mathcal{M}$ denotes the configuration space of the system (e.g., parameter space or an extended structural space). Any training procedure defines an update rule of the form
\[
x_{t+1} = \Phi(x_t),
\]
where $\Phi$ represents the learning algorithm. In the continuous-time limit, this discrete evolution can be written as
\[
\dot{x} = F(x),
\]
where $F(x)$ defines a vector field on $\mathcal{M}$. The trajectory $x(t)$ therefore describes how the system reorganizes its internal configuration over training time.
This formulation applies broadly across learning paradigms. Gradient descent, policy optimization, evolutionary strategies, and other adaptive procedures all generate flows in configuration space determined by the structure of $F$.

The central question of this work concerns the structural properties of the learning dynamics itself. In particular, we ask: what classes of vector fields can arise as learning rules, and how do their structural differences affect the behavior of learning systems? Learning rules can in principle be classified in many different ways. One may examine properties such as symmetry, conservation laws, divergence structure, or time-reversal behavior. These perspectives reveal different aspects of the dynamics.

Among these possibilities, however, one distinction is particularly fundamental. A dynamical system may or may not admit a global scalar ordering principle. In other words, the governing vector field may be reducible to the gradient of a scalar potential. When such a representation exists, the dynamics are globally constrained by the monotonic decrease of the scalar potential. When it does not, the system can exhibit intrinsically non-potential behavior that cannot be captured by scalar optimization. This distinction provides a natural structural classification of learning rules.

\begin{figure}[ht]
    \centering
    \includegraphics[width=\linewidth]{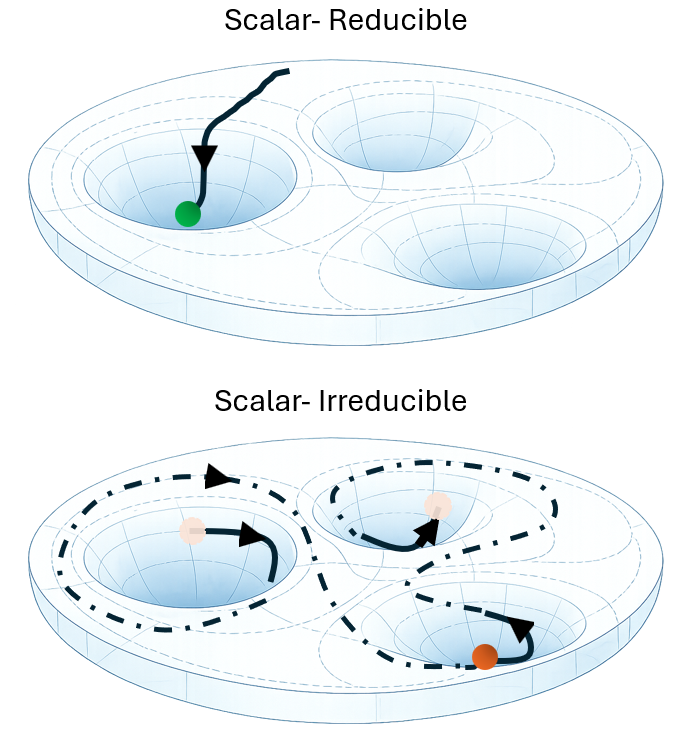}
    \caption{\textbf{Conceptual illustration of scalar-reducible and scalar-irreducible learning dynamics.}  In scalar-reducible dynamics (top), the flow follows the gradient of a scalar objective and typically converges into a single basin, locking the system into one regime. In scalar-irreducible dynamics (bottom ), rotational components of the dynamics allow trajectories to repeatedly traverse different regions of the landscape, enabling endogenous regime switching.}
    \label{model}
\end{figure}

\subsection{Scalar-Reducible Dynamics}

We say that a dynamical system is \textbf{scalar-reducible} if there exists a continuously differentiable scalar function $V : \mathcal{M} \to \mathbb{R}$ such that
\[
\frac{d}{dt} V(x(t)) = \nabla V(x) \cdot F(x) \le 0
\quad \text{for all } x \in \mathcal{M}.
\]
In this case, $V$ acts as a Lyapunov function that monotonically decreases (or increases) along trajectories, so the dynamics are globally ordered by a scalar quantity. The evolution of the system can therefore be interpreted as moving consistently toward lower values of the scalar potential $V$.

A useful way to understand scalar-reducible dynamics is through a Helmholtz-type decomposition of the vector field. In Euclidean domains, sufficiently regular vector fields can be written as
\[
F(x) = -\nabla V(x) + R(x),
\]
where $-\nabla V(x)$ represents the gradient component and $R(x)$ is a rotational component. 

A special case of scalar-reducible dynamics is gradient flow, where the rotational component vanishes, $R(x) = 0$, so that
\[
F(x) = -\nabla V(x). 
\]
More generally, even when a rotational component is present, the system may still remain scalar-reducible. In particular, if the rotational component is orthogonal to the gradient of $V$,
\[
\nabla V(x) \cdot R(x) = 0,
\]
then the scalar function $V$ still decreases monotonically along trajectories:
\[
\dot V(x) = \nabla V(x) \cdot F(x) = 
\]
\[
\nabla V(x)\cdot(-\nabla V(x) + R(x))
= -|\nabla V(x)|^2 < 0.
\]
Thus the system remains scalar-reducible even though the vector field contains rotational structure. In this case the trajectories may spiral within level sets of $V$ while still descending the scalar potential.

The existence of such a global scalar ordering implies several structural properties:
\begin{itemize}
    \item \textbf{Trajectory Ordering}: All trajectories are partially ordered by the scalar $V$.
    \item \textbf{Absence of Cyclic Recurrence}: Closed orbits are excluded except on measure-zero invariant sets.
    \item \textbf{Asymptotic Convergence}: Under mild compactness conditions, trajectories approach invariant sets contained within level sets of $V$.
    \item \textbf{Path-Independent Evaluation}: The “quality” of a state can be evaluated independent of the trajectory taken to reach it.
\end{itemize}

\subsection{Scalar-Irreducible Dynamics}

We define a system as \textbf{scalar-irreducible} if no global continuously differentiable scalar function $V$ exists such that
\[
\nabla V(x) \cdot F(x) \le 0
\quad \text{for all } x \in \mathcal{M}.
\]
Equivalently, the flow does not admit a global Lyapunov ordering. In scalar-irreducible systems:
\begin{itemize}
    \item No global ranking of states exists.
    \item Trajectories may exhibit cyclic recurrence or persistent non-convergent behavior.
    \item Path dependence may be intrinsic.
    \item Structural transitions may occur that cannot be captured by monotonic descent.
\end{itemize}

\section{Modern ML as Scalar-Reducible Learning}

We now examine where contemporary machine learning systems lie within the structural classification introduced above. Rather than surveying architectures individually, we analyze the dominant training paradigm underlying modern ML.

Our central observation is that modern ML systems overwhelmingly belong to the class of \emph{scalar-reducible dynamics}. Regardless of differences in architecture, training protocol, or learning paradigm, the parameter evolution of these systems is typically governed by the monotonic variation of a scalar functional. Importantly, this conclusion does not depend on whether the system is formulated with an explicit loss function or a clearly defined task objective. In many cases—such as variational inference, energy-based models, or biologically inspired learning rules—the objective may appear implicit or absent at the algorithmic level. Nevertheless, the resulting parameter dynamics can still be expressed in terms of a scalar quantity whose variation globally orders the system’s evolution.

Thus the defining structural property of modern machine learning is not the presence of supervision, labels, or explicit objectives, but the reducibility of its learning dynamics to a global scalar ordering.

\subsection{The Structural Core of Modern Training}

Across supervised learning~\cite{Kotsiantis,Singh2016,Nasteski2017,Jiang2020}, reinforcement learning~\cite{Jia2020,Shakya2023}, self-supervised learning~\cite{Hastie2013,Rani2023,Shukla2025}, energy-based models~\cite{Linsker1988,Rao1999,Jaynes1957,Hinton2002}, and variational inference, parameter updates take the general form
\[
\theta_{t+1} = \theta_t + \Delta \theta_t,
\]
where $\Delta \theta_t$ is derived—explicitly or implicitly—from a scalar objective functional $\mathcal{L}(\theta)$.

In the continuous-time limit, this induces dynamics
\[
\dot{\theta} = -G(\theta)\nabla \mathcal{L}(\theta),
\]
where $G(\theta)$ is a positive semi-definite metric tensor or preconditioning operator (e.g., identity for SGD, adaptive diagonal matrices for Adam, Fisher metric for natural gradient). Mathematically, $G(\theta)$ need not be positive semi-definite in general dynamical systems. However, in optimization-based learning algorithms it is typically chosen to be positive semi-definite, so that the update direction constitutes a descent direction for the objective $\mathcal{L}$.

This class of systems admits a global Lyapunov function:
\[
\frac{d}{dt} \mathcal{L}(\theta(t))
= \nabla \mathcal{L}(\theta) \cdot \dot{\theta}
= - \nabla \mathcal{L}(\theta)^\top G(\theta) \nabla \mathcal{L}(\theta)
\le 0.
\]
Thus, the training dynamics are scalar-reducible:
the scalar functional $\mathcal{L}$ monotonically decreases along trajectories. Importantly, the presence of stochasticity (e.g., minibatch noise) does not alter this structural property; it perturbs trajectories but does not eliminate the existence of the underlying scalar ordering.

\subsection{Beyond Explicit Loss Functions}

One might object that not all learning paradigms explicitly specify a task-level loss function. For example, several frameworks—including variational inference, predictive coding, and energy-based models—are often described as learning without a predefined objective in the conventional supervised-learning sense. However, even in these cases a scalar quantity typically governs the learning dynamics at a deeper level. Rather than directly minimizing a task loss, the system adjusts its parameters to improve the consistency between an internal probabilistic model and observed data.

A canonical example arises in variational inference.
Here the goal is to approximate an intractable posterior distribution $p(z \mid x)$ by introducing a parameterized distribution $q_\theta(z)$. Learning proceeds by reducing the discrepancy between the two distributions. This discrepancy is commonly measured by the Kullback–Leibler divergence, leading to the variational objective
\[
\mathcal{F}(\theta)
=
\mathrm{KL}\!\left(q_\theta(z) \,\|\, p(z \mid x)\right).
\]
Gradient-based learning then takes the general form
\[
\dot{\theta}
=
- G(\theta)\nabla_\theta \mathcal{F}(\theta),
\]
where $G(\theta)$ is a positive semi-definite metric or preconditioning operator.
Applying the chain rule gives
\[
\frac{d}{dt}\mathcal{F}(\theta(t))
=
\nabla_\theta \mathcal{F}(\theta)\cdot\dot{\theta}
=
-
\nabla_\theta \mathcal{F}(\theta)^\top
G(\theta)
\nabla_\theta \mathcal{F}(\theta)
\le 0,
\]
which shows that the scalar function $\mathcal{F}$ monotonically decreases along trajectories whenever $G(\theta)$ is positive semi-definite.

Thus, even in learning paradigms that are often described as objective-free at the task level, the parameter dynamics remain governed by a scalar function. The evolution of the system is therefore scalar-reducible.

\subsection{Implicit Optimization Systems}

Certain adaptive systems, such as Hebbian learning~\cite{Oja1982}, spike-timing-dependent plasticity (STDP), and self-organizing maps~\cite{Kohonen1982}, do not explicitly specify a global objective. Instead, they rely on biologically inspired learning rules in which parameter updates are defined through local interactions. However, many such local learning rules can be shown to admit equivalent Lyapunov or energy-function formulations. 

A canonical example is Oja's rule, a normalized form of Hebbian learning. In this system the update rule is defined purely through local interactions between the input and the neuron output, without reference to any global objective.

Let $x \in \mathbb{R}^n$ denote the input vector and $y = w^\top x$ the neuron output. The learning rule specifies the weight dynamics
\[
\dot w = yx - y^2 w .
\]
Taking expectation over the input distribution gives the mean dynamics
\[
\dot w
=
Cw - (w^\top C w)w,
\]
where $C = \mathbb{E}[xx^\top]$ is the covariance matrix. This update rule is the learning mechanism itself and does not explicitly derive from any optimization objective. 

We now show that despite the absence of an explicitly defined objective, the resulting dynamics admit a scalar Lyapunov function. Consider the scalar function
\[
V(w) = -\frac12 w^\top C w.
\]
The gradient is
\[
\nabla V = -Cw.
\]
The time derivative along trajectories of the learning dynamics is
\[
\dot V
=
\nabla V \cdot \dot w
=
(-Cw)^\top
(Cw - (w^\top C w)w).
\]
Expanding gives
\[
\dot V
=
- w^\top C^2 w
+
(w^\top C w)^2
\le 0.
\]
Thus $V(w)$ acts as a Lyapunov function for the system. Although the learning rule is defined purely through local interactions, the parameter dynamics remain governed by a scalar ordering principle, and the system converges to the principal eigenvector of $C$.

\subsection{Adversarial and Game-Theoretic Learning}

Adversarial learning systems, such as generative adversarial networks (GANs)~\cite{goodfellow2014} or multi-agent reinforcement learning, involve coupled parameter dynamics between multiple agents. These systems may exhibit local rotational components in joint parameter space. However, they are still typically derived from minimax optimization of a scalar functional $\mathcal{L}(\theta_1,\theta_2)$.

A minimal example is the bilinear minimax game
\[
\min_x \max_y \mathcal{L}(x,y),
\qquad
\mathcal{L}(x,y)=xy .
\]
Gradient dynamics take the form
\[
\dot x = -\frac{\partial \mathcal{L}}{\partial x},
\qquad
\dot y = \frac{\partial \mathcal{L}}{\partial y}.
\]
This yields
\[
\dot x = -y,
\qquad
\dot y = x.
\]
The joint dynamics can be written as
\[
\begin{pmatrix}
\dot x\\
\dot y
\end{pmatrix}
=
\begin{pmatrix}
0 & -1\\
1 & 0
\end{pmatrix}
\begin{pmatrix}
x\\
y
\end{pmatrix}.
\]
The resulting trajectories satisfy
\[
x^2 + y^2 = \text{const},
\]
which correspond to circular motion in parameter space. Thus the system exhibits persistent rotational dynamics.

Importantly, this rotational behavior arises despite the fact that the system is derived directly from a scalar functional $\mathcal{L}(x,y)$. The rotation reflects the saddle geometry of the minimax objective rather than a fundamentally scalar-irreducible structure.

\begin{table*}[t]
\centering
\caption{Classification of learning dynamics in existing machine learning frameworks. 
Despite their apparent diversity, most approaches can be reduced to dynamics driven by a scalar objective. 
The framework proposed in this work instead introduces scalar-irreducible dynamics that enable endogenous regime switching.}

\begin{tabular}{l l l l l}
\toprule
\textbf{Framework} & \textbf{Objective} & \textbf{Dynamics} & \textbf{Mathematical Form} & \textbf{Regime Switching} \\
\midrule

Supervised learning 
& Explicit loss 
& Gradient descent 
& $\dot{\theta} = -\nabla V(\theta)$ 
& External \\

Reinforcement learning 
& \parbox[t]{3cm}{\raggedright Reward maximization  \\[4pt]} 
& \parbox[t]{3.5cm}{\raggedright Policy gradient (value updates)  \\[4pt]}
& $\dot{\theta} = \nabla J(\theta)$ 
& External \\

\parbox[t]{3.5cm}{\raggedright Rule-based adaptive learning \\[4pt]} 
& Implicit objective 
& Local update rules 
& \parbox[t]{3.5cm}{\raggedright Effective gradient dynamics \\[4pt]}   
& External \\

Multi-agent learning 
& Game payoff 
& Coupled gradient
& \parbox[t]{3.5cm}{\raggedright Gradient game dynamics  \\[4pt]}    
& External \\

\parbox[t]{3.5cm}{\raggedright \textbf{Scalar-irreducible learning (this work)} \\[4pt]}
& None 
& \parbox[t]{3.5cm}{\raggedright Intrinsic vector-field dynamics \\[4pt]}  
& $\dot{\theta} = F(\theta),\; \nexists V$ 
& \textbf{Endogenous} \\

\bottomrule
\end{tabular}
\label{tab:learning_dynamics}
\end{table*}

\section{Endogenous Regime-Switching Learning by Scalar-Irreducible Dynamics}

The discussion above raises a natural question: if modern learning systems are predominantly scalar-reducible, what role—if any—could scalar-irreducible dynamics play in learning?

A central aspect of adaptive learning is the ability of a system to reorganize its internal dynamical regime when the current mode of operation becomes inadequate. Such reorganization often appears as a transition between attractors or dynamical regimes of the system.

Attractor or regime transitions by themselves do not require scalar irreducibility. In practice, such transitions can arise through several mechanisms. Stochastic perturbations may drive escape from one basin of attraction to another. Time-dependent objectives or curricula may reshape the effective landscape over training time. External interventions, such as resets or architectural modifications, can also induce transitions between dynamical regimes. In each of these cases, however, the driving force for the transition originates from external sources rather than from the autonomous dynamics of the system itself.

For a truly autonomous learning system, the underlying dynamical rules remain fixed. The system must determine, through its own internal dynamics, when a currently stable regime has become inadequate and initiate structural reorganization accordingly. In this setting, regime transitions must arise endogenously rather than through externally prescribed schedules.

In the following, we demonstrate that if regime switching is required to occur endogenously within a system governed by fixed internal dynamics—without stochastic escape, externally time-dependent objectives, or external intervention—then the resulting dynamics cannot, in general, be represented as the monotonic evolution of a single global scalar functional. This observation motivates the investigation of scalar-irreducible dynamical structures as a natural framework for autonomous regime-regulating learning systems.

\subsection{Endogenous repeated regime-switching learning}

We consider an autonomous dynamical system on an extended state space
\[
\dot z = F(z), \qquad z \in K,
\]
where \(K\subset \mathbb{R}^n\) is a compact positively invariant set and \(F\in C^1(K)\). The state \(z\) may include both fast configuration variables and slower internal structural or viability variables. The compactness of \(K\) is a standard boundedness assumption ensuring that the long-time dynamics remain well defined.

We assume that the state space contains a collection of pairwise disjoint subsets
\[
\Omega_1,\Omega_2,\dots,\Omega_m \subset K,
\]
which represent distinct dynamical regimes. These may correspond to attractor neighborhoods, metastable organizations, or dynamically coherent modes of operation. We also introduce a transition region
\[
T \subset K,
\]
consisting of states in which the system is actively reorganizing between such regimes.

The object of interest is not a single transient escape event, but a learning process in which the system retains the capacity to leave a currently occupied regime and reorganize repeatedly over long time horizons.

\textbf{Endogenous repeated regime-switching learning.}
An autonomous system \(\dot z = F(z)\) is said to exhibit \emph{endogenous repeated regime-switching learning} if there exists a trajectory \(z(t)\) and a sequence of times
\[
0 < t_1^- < t_1^+ < t_2^- < t_2^+ < \cdots, \qquad t_k^\pm \to \infty,
\]
such that for each \(k\):
\begin{enumerate}
    \item there exists some regime region \(\Omega_{i_k}\) in which the trajectory dwells for a nontrivial time interval before \(t_k^-\);
    \item the trajectory passes through the transition region \(T\) on \([t_k^-,t_k^+]\);
    \item after the transition, the trajectory enters a regime region \(\Omega_{j_k}\) with \(j_k \neq i_k\);
    \item the transition is generated solely by the autonomous internal dynamics \(F\), with no external forcing, no externally time-dependent objective, and no stochastic escape term.
\end{enumerate}

This definition isolates the specific learning behavior of interest: not mere convergence, not one-shot escape, but sustained internally generated reorganization of dynamical regime.

\subsection{Limitation of scalar-reducible dynamics}

We first recall the structural consequence of scalar reducibility. Suppose the dynamics are scalar-reducible, so that there exists a continuously differentiable scalar function
\[
V:K \to \mathbb{R}
\]
satisfying
\[
\dot V(z) = \nabla V(z)\cdot F(z) \le 0
\qquad \text{for all } z\in K.
\]
Along any trajectory \(z(t)\), the quantity \(V(z(t))\) is therefore nonincreasing. Because \(K\) is compact and \(V\) is continuous, \(V\) is bounded below on \(K\), and thus \(V(z(t))\) must converge to a finite limit as \(t\to\infty\).

A standard consequence (LaSalle invariance principle) is that the long-time dynamics are confined to the invariant set
\[
E=\{z\in K:\dot V(z)=0\}.
\]
In other words, although the system may transiently explore different regions of state space, its asymptotic behavior must eventually lie within the zero-dissipation manifold where the scalar descent halts.

This observation already imposes an important structural restriction. Scalar-reducible dynamics admit a global ordering: trajectories may descend through the state space, but their long-time organization is compressed by a single scalar quantity. Multiple attractors are not forbidden, but the asymptotic dynamics cannot indefinitely sustain large-scale reorganization unless such motion occurs entirely within the zero-dissipation set \(E\).

To sharpen this observation, consider a regime-switching event in which the system leaves one dynamical organization and actively reorganizes into another. If such a transition necessarily involves passing through a region \(T\) where the scalar dissipation is strictly negative,
\[
\dot V(z)\le -\eta <0,
\]
then each transition reduces \(V\) by a finite amount. Indeed, if the system spends at least a time \(\tau_0\) in this transition region, then
\[
\Delta V
=
\int_{t}^{t+\tau_0}\dot V(z(t'))\,dt'
\le
-\eta\tau_0 .
\]

Since \(V\) is bounded below on the compact state space \(K\), only finitely many such nondegenerate transition events can occur. Repeated regime reorganization therefore cannot persist indefinitely under globally scalar-descending dynamics unless the transitions themselves become asymptotically degenerate.

The essential point is therefore not the existence of multiple attractors, but the impossibility of sustaining repeated nondegenerate endogenous reorganization under a globally monotone scalar ordering.

\subsection{Implication for autonomous learning}

The argument above has a direct implication for autonomous learning dynamics. Autonomous learning systems of the type defined above require more than one-way descent. They must retain the capacity to dwell, destabilize, reorganize, recover, and reorganize again under fixed internal dynamical rules. This requires a dynamical structure that is not globally compressed by a single scalar ordering.

In scalar-reducible systems, the dynamics are globally ordered. The system may approach different attractors, and transient reorganizations may occur, but any nondegenerate transition necessarily consumes scalar potential. If this potential is bounded below, repeated endogenous reorganization must eventually terminate or degenerate into asymptotically vanishing motion. Autonomous learning understood as sustained endogenous regime reorganization therefore cannot be realized within strictly scalar-reducible flows.

Scalar-irreducible dynamics remove this restriction. Without a global scalar ordering, the dynamics are not constrained to asymptotically exhaust a bounded potential, allowing recurrent internally generated regime switching under fixed internal rules.

This observation motivates the constructions developed below, where learning architectures are built directly from scalar-irreducible dynamical flows.

\section{Minimal Dynamical Model for Scalar-Irreducible Learning}

Based on the theoretical considerations above, we construct a minimal dynamical system exhibiting endogenous repeated regime switching driven by scalar-irreducible dynamics.

Importantly, scalar-irreducible dynamics do not imply the absence of gradient-driven components. A gradient (scalar-reducible) contribution can still play an essential stabilizing role in the structural evolution, preventing uncontrolled drift of parameters. What distinguishes the present framework is the presence of an additional irreducible component of the learning flow that cannot be represented as the gradient of any scalar potential. It is this irreducible component that enables the system to explore parameter space and repeatedly cross dynamical regime boundaries.

We therefore consider a dynamical system composed of two coupled layers evolving on separated timescales: a fast dynamical layer describing the system's internal state evolution, and a slow structural layer describing adaptive plasticity of internal parameters. The slow layer is activated only when the fast dynamics exhibit pathological dynamical signatures.

\subsection{Fast dynamical layer}

The fast subsystem evolves according to a nonlinear dynamical system
\[
\dot{\mathbf{x}} = \mathbf{F}(\mathbf{x};\boldsymbol{\theta}),
\]
where
$\mathbf{x}(t)\in\mathbb{R}^n$ denotes the fast dynamical state and $\boldsymbol{\theta}(t)\in\mathbb{R}^m$ represents slowly varying structural parameters.

For concreteness we adopt a minimal two-dimensional excitable dynamics (FitzHugh–Nagumo–type system)
\[
\dot u = u - \frac{u^3}{3} - v + \theta_1
\]
\[
\dot v = \varepsilon (u + a - b v)
\]
where
\[
\mathbf{x}=(u,v),\qquad \boldsymbol{\theta}=(\theta_1,\theta_2)
\]
and $0<\varepsilon\ll1$ ensures separation of timescales.

Depending on parameter values, the system exhibits qualitatively distinct dynamical characters such as: quiescent fixed-point dynamics, excitable responses, and oscillatory limit-cycle behavior. Thus the fast layer naturally possesses multiple dynamical regimes separated by bifurcation boundaries in parameter space.

\subsection{Dynamical evaluation of internal performance}

The system does not optimize an externally specified objective function.
Instead, it evaluates the health of its own internal dynamics. We define a set of dynamical indicators
\[
\mathbf{M}(t)
=
(M_{\mathrm{freeze}},
M_{\mathrm{cycle}},
M_{\mathrm{mono}})
\]
representing freezing or dynamical inactivity, pathological cyclic trapping, and long-term dynamical monotony.  

The freezing indicator measures the suppression of dynamical activity,
while the cyclic indicator detects trapping in low-dimensional periodic motion. To quantify long-term diversity of the trajectory, we consider the variance of the state over a sliding window
\[
R(t)=\mathrm{Var}_{t-T_R}^{t}\big(\mathbf{x}\big).
\]
Small values of $R(t)$ indicate that the system remains confined to a narrow dynamical pattern over extended periods.
We therefore define a monotony indicator
\[
M_{\mathrm{mono}} = \exp(-\gamma R(t)),
\]
which increases when the trajectory lacks long-term diversity.

From these quantities we construct a scalar badness functional
\[
B(t)
=
w_f M_{\mathrm{freeze}}
+
w_c M_{\mathrm{cycle}}
+
w_m M_{\mathrm{mono}},
\]
where $w_f$, $w_c$, and $w_m$ are weighting coefficients. Badness therefore measures pathological dynamical behavior of the system itself rather than task performance.

To avoid transient fluctuations, the system evaluates a smoothed stress variable
\[
\dot S = \frac{1}{\tau_s}\big(B(t)-S\big),
\]
which represents the accumulated dynamical stress generated by sustained unhealthy dynamical patterns.

\begin{figure*}[ht]
    \centering
    \includegraphics[width=\linewidth]{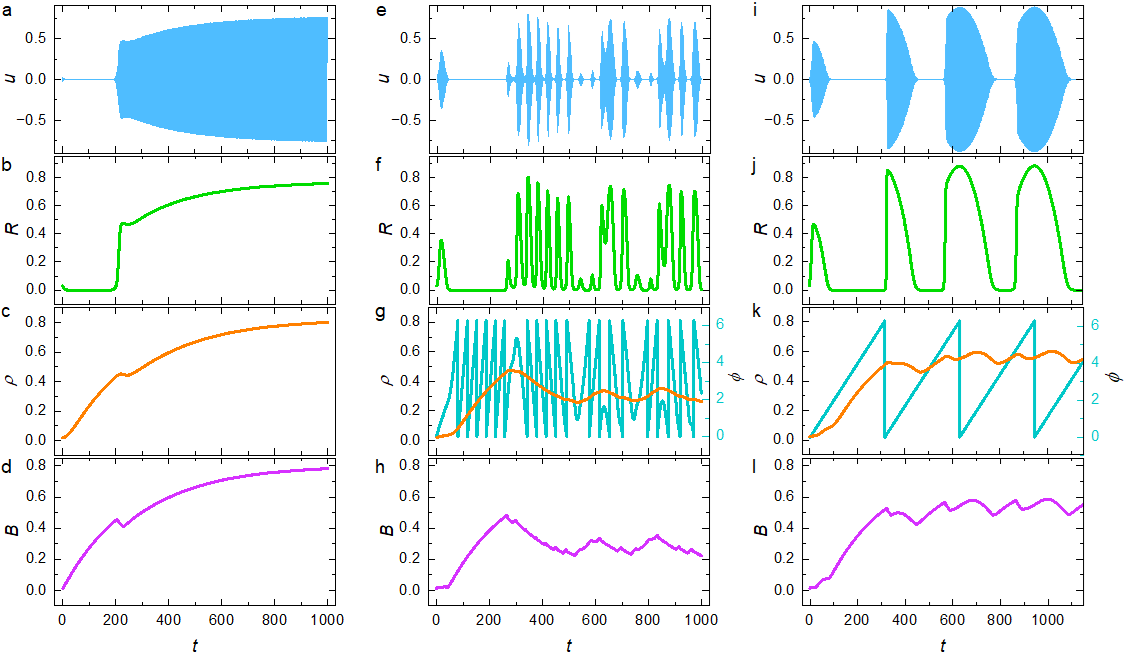}
    \caption{\textbf{Comparison between scalar-reducible and scalar-irreducible learning dynamics.} The left panel (a–d) shows the scalar-reducible baseline, the middle panel (e–h) corresponds to the scalar-irreducible system, and the right panel (i-l) shows the externally swept case. From top to bottom the figures display the fast dynamical variable $u$, dynamical activity $R$ of $u$, the slow structural variables ($\rho$ and $\phi$), and the dynamical badness $B$.}
    \label{model}
\end{figure*}

\subsection{Slow plasticity dynamics}

Structural parameters evolve on a slow timescale
\[
\dot{\boldsymbol{\theta}}=\mathbf{G}(\boldsymbol{\theta},S)
\]
where plasticity is activated only when stress exceeds a threshold $S_c$
\[
\dot{\boldsymbol{\theta}} =
H(S-S_c), \mathbf{G}(\boldsymbol{\theta})
\]
with $H$ the Heaviside function. Thus plasticity is stress-gated rather than continuously active.

In conventional optimization-based learning systems, structural change is driven by the gradient of a scalar objective function
\[
\dot{\boldsymbol{\theta}}
=
-\eta\nabla_{\boldsymbol{\theta}} L(\boldsymbol{\theta})
\]
for some scalar loss $L$.

Such dynamics are scalar-reducible, because the entire structural evolution is determined by a single scalar potential.
The flow therefore satisfies
\[
\nabla\times \dot{\boldsymbol{\theta}} = 0
\]
and the dynamics cannot sustain rotational motion in parameter space.

In contrast, we introduce irreducible plasticity dynamics
\[
\dot{\boldsymbol{\theta}}
=
H(S-S_c)\Big[
-\eta\nabla U(\boldsymbol{\theta})
+
\mathbf{R}(\boldsymbol{\theta})
\Big]
\]
where
$U(\boldsymbol{\theta})$ is an optional stabilizing potential and
$\mathbf{R}(\boldsymbol{\theta})$ is a curl component
\[
\nabla\times \mathbf{R}(\boldsymbol{\theta})\neq0 .
\]

The presence of this rotational component implies that
\[
\dot{\boldsymbol{\theta}}
\neq
\nabla L
\]
for any scalar $L$. The dynamics therefore become scalar-irreducible.

A minimal realization in two dimensions is
\[
\mathbf{R}(\boldsymbol{\theta})
=
\omega
\begin{pmatrix}
-\theta_2\
\theta_1
\end{pmatrix}
\]
which produces rotational flow in parameter space.

\subsection{Mechanism for regime switching}

Because the fast dynamics depend on structural parameters
\[
\dot{\mathbf{x}} = \mathbf{F}(\mathbf{x};\boldsymbol{\theta}),
\]
movement of $\boldsymbol{\theta}$ across bifurcation boundaries can change the qualitative character of the fast dynamics.

The system therefore exhibits the following closed feedback loop:
1. Fast dynamics generate dynamical statistics
2. Pathological behavior increases badness
3. Badness accumulates stress
4. Stress activates slow plasticity
5. Scalar-irreducible flow moves parameters in a non-gradient direction
6. Structural parameters cross dynamical bifurcation boundaries
7. A new dynamical regime emerges.

Thus regime switching is not imposed externally but arises from the internal coupling between dynamical evaluation and irreducible structural adaptation.

\subsection{Results and Discussion}

Our results are summarized in Fig.~\ref{model}, which compares the dynamical behavior of the scalar-reducible baseline and the scalar-irreducible learning system. Shown are the fast dynamical variable $u$, the dynamical activity $R$ of $u$, the slow structural variables $\rho$ and $\phi$, and the badness variable $B$. In the scalar-reducible case (Fig.~\ref{model}a–d), the system initially undergoes a transient regime transition. However, the slow structural variable quickly converges to a stationary value. Once this occurs, structural adaptation effectively ceases and the system becomes trapped in a single dynamical regime, characterized by persistent oscillations of the fast variable while the structural variables remain frozen. Once this trapping occurs, no further regime switching takes place. Correspondingly, the badness $B$ remains relatively high and gradually saturates.

In contrast, the scalar-irreducible system (Fig.~~\ref{model}e–h) exhibits persistent regime switching. The fast dynamics repeatedly alternate between quiescent and oscillatory states, producing frequent transitions in the fast variable $u$ and the dynamical activity $R$. These transitions are intermittent rather than periodic: both $u$ and $R$ show irregular bursts, while the slow structural variables evolve in a feedback-regulated manner.
At the same time, the badness $B$ is maintained at a lower level than in the reducible baseline, indicating that repeated endogenous switching improves the overall dynamical health of the system.

For comparison, we also consider an externally swept case (Fig.~\ref{model}i–l), in which $\phi$ is not regulated by badness but is instead driven at a constant rate. This case also exhibits repeated regime transitions. However, the resulting switching pattern is much more regular and closely tracks the imposed evolution of $\phi$. This demonstrates that repeated switching alone is not sufficient to identify endogenous learning dynamics, since externally imposed parameter scanning can produce switching as well. What distinguishes the scalar-irreducible system is that its switching is generated internally through the closed feedback loop linking fast dynamics, badness evaluation, and slow structural adaptation, rather than by an externally prescribed schedule.


\section{Conclusions}

In modern machine learning, learning dynamics are typically organized around optimization of a predefined objective. While highly effective, such dynamics tend to confine the system to a single regime unless transitions are externally imposed through mechanisms such as scheduling, noise injection, or restarts. The framework proposed here suggests a different possibility: learning dynamics that can internally generate regime transitions through their own intrinsic structure. Such dynamics may provide a new approach to addressing one of the persistent difficulties in training complex systems—the tendency to become trapped within a particular regime of behavior.

More broadly, the implications extend beyond practical training considerations. For autonomous learning systems, the ability to transition between regimes without external intervention may represent a fundamental threshold. A system that can internally regulate when to remain within a regime and when to enter a different one gains the capacity to reorganize how it explores and adapts. In this sense, endogenous regime switching may constitute an early step toward a form of intelligence in which the direction of exploration is determined from within the dynamics of the system itself rather than prescribed from the outside.

\bibliographystyle{unsrt}
\bibliography{scalar}

\clearpage

\end{document}